\documentclass{sig-alternate-05-2015}

\usepackage{cite}
\usepackage[latin9]{inputenc}

\usepackage{amsthm}

\usepackage{amsmath}
\usepackage{amssymb}
\usepackage{enumerate}
\usepackage{centernot}
\usepackage{multicol,comment}
\usepackage{xspace}
\usepackage{enumitem}
\usepackage{environ}
\usepackage{times}
\usepackage{float}
\usepackage{algorithmicx}
\usepackage{algpseudocode}
\usepackage{bm}
\usepackage{lscape}
\usepackage{pdflscape}
\usepackage{wrapfig}
\usepackage{rotating}
\usepackage{epstopdf}
\usepackage{setspace}
\usepackage{breqn}

\usepackage{epsfig}
\usepackage{xcolor}

\usepackage{algorithm}

\usepackage[hyphens]{url}

\usepackage{graphicx,subcaption}

\usepackage[unicode,pdfstartview=FitH]{hyperref}

\algnewcommand\algorithmicinput{\textbf{Input:}}
\algnewcommand\INPUT{\item[\algorithmicinput]}
\algnewcommand\algorithmicoutput{\textbf{Output:}}
\algnewcommand\OUTPUT{\item[\algorithmicoutput]}
\algnewcommand\algorithmicproc{\textbf{Procedure}}
\algnewcommand\Proc{\item[\algorithmicproc]}

\DeclareMathOperator*{\argmin}{\mathrm{argmin}}
\DeclareMathOperator*{\argmax}{\mathrm{argmax}}

\makeatletter
\def\@copyrightspace{\relax}
\makeatother

\newcommand*{\Resize}[2]{\resizebox{#1}{!}{$#2$}}%

\newcommand{\lyxmathsym}[1]{\ifmmode\begingroup\def\b@ld{bold}
	\text{\ifx\math@version\b@ld\bfseries\fi#1}\endgroup\else#1\fi}
\newcommand{\nl}{\textsf{NULL}\xspace}

\begin{document}


\title{Blocking Transferability of Adversarial Examples in Black-Box Learning Systems}

\numberofauthors{5}
\author{
	\alignauthor Hossein Hosseini\\
	\email{hosseinh@uw.edu}
	\alignauthor Yize Chen\\
	\email{yizechen@uw.edu}
	\alignauthor Sreeram Kannan\\
	\email{ksreeram@uw.edu}
	\and  
	\alignauthor Baosen Zhang\\
	\email{zhangbao@uw.edu}
	\alignauthor Radha Poovendran\\
	\email{rp3@uw.edu}\vspace{0.15cm}
	\and  
	\affaddr{Department of Electrical Engineering, University of Washington, Seattle, WA}\vspace{0.3cm}
}

\maketitle


\begin{abstract}
Advances in Machine Learning (ML) have led to its adoption as an integral component in many applications, including banking, medical diagnosis, and driverless cars. To further broaden the use of ML models, cloud-based services offered by Microsoft, Amazon, Google, and others have developed ML-as-a-service tools as black-box systems. However, ML classifiers are vulnerable to adversarial examples: inputs that are maliciously modified can cause the classifier to provide adversary-desired outputs. Moreover, it is known that adversarial examples generated on one classifier are likely to cause another classifier to make the same mistake, even if the classifiers have different architectures or are trained on disjoint datasets. This property, which is known as {\it transferability}, opens up the possibility of attacking black-box systems by generating adversarial examples on a substitute classifier and transferring the examples to the target classifier. 

Therefore, the key to protect black-box learning systems against the adversarial examples is to block their transferability. To this end, we propose a training method that, as the input is more perturbed, the classifier smoothly outputs lower confidence on the original label and instead predicts that the input is ``invalid''. In essence, we augment the output class set with a \nl label and train the classifier to reject the adversarial examples by classifying them as \nl. In experiments, we apply a wide range of attacks based on adversarial examples on the black-box systems. We show that a classifier trained with the proposed method effectively resists against the adversarial examples, while maintaining the accuracy on clean data. 

\end{abstract}

\section{Introduction}

Machine learning (ML) techniques have substantially advanced in the past decade and are having significant impacts on everyday lives. Success of ML algorithms has led to an explosion in demand. ML models are also increasingly applied in security-sensitive and critical systems; for instance, autonomous vehicles heavily rely on computer vision algorithms \cite{cirecsan2012multi} and the system security is of paramount importance. To further broaden and simplify the use of ML algorithms, cloud-based services offered by Amazon, Google, Microsoft, BigML, and others have developed ML-as-a-service tools. Thus, users and companies can readily benefit from ML applications without having to train or host their own models. 
The wide deployment of the remotely-hosted systems signifies the importance of studying the security of the {\it black-box} learning systems, where the adversary does not have access to the model parameters.

\begin{figure}[t]
	\centering
	\includegraphics[width=0.475\textwidth]{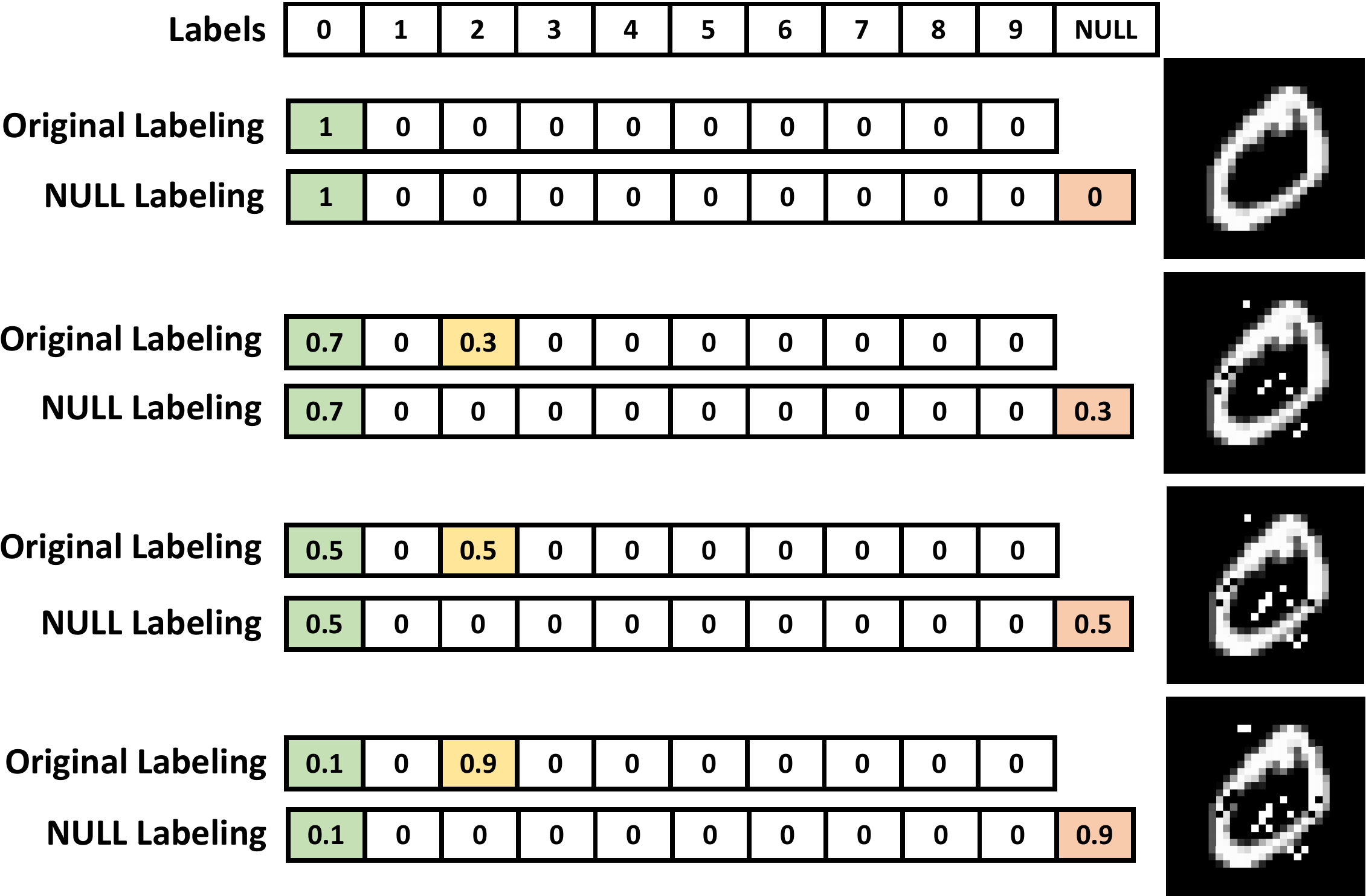}
	\caption{Illustration of the original and \nl labeling methods on an image from MNIST dataset, and three adversarial examples with different perturbations. 
The classifier assigns a probability vector to each image. With more perturbation, the \nl labeling method assigns higher probability to the \nl label, while the original labeling increases the probabilities of other labels.} 
	\label{fig:images}
\end{figure}

The implicit assumption in training ML models is that they will be deployed in benign settings. However, many works have pointed out their vulnerability in adversarial scenarios \cite{barreno2006can,barreno2010security,huang2011adversarial,biggio2014security,amodei2016concrete,papernot2016towards}. One type of the vulnerabilities of the ML classifiers is that an adversary can significantly change the algorithm output by slightly perturbing the input. Such modified inputs are called {\it adversarial examples}~\cite{szegedy2013intriguing}. The adversary's goal is to cause the model to either misclassify the input (misclassification attack) or output a target label (targeted attack). 
Attacks based on adversarial examples can be categorized according to the adversary's knowledge about the system. If the adversary has full knowledge about the internal structure of the classifier, she can solve an optimization problem to obtain the smallest change that has the largest adversary-desired influence on the output~\cite{goodfellow2014explaining,carlini2016towards,papernot2016limitations}. 

However, in many real-world applications, the adversary may not have access to the system parameters. Therefore, a more realistic attack model is the black-box model, where we assume that the adversary only possesses a small dataset and has an {\it oracle} access to the system, i.e., she can query the classifier with any input and obtain its output label~\cite{papernot2016practical}. Because of the adversary's limited knowledge, it may seem that black-box systems are secure against adversarial examples. However, it has been shown that adversarial examples, generated on one classifier, are likely to cause another classifier to make the same mistake, even if the the classifiers have different architectures or are trained on disjoint datasets. This property is known as {\it transferability}~\cite{szegedy2013intriguing}. In~\cite{papernot2016practical,papernot2016transferability}, Papernot et al. used the transferability property to attack remotely-hosted learning systems~(e.g. Amazon AWS) using the adversarial examples generated on a substitute classifier.

{\it Therefore, the key to protect black-box learning systems against the adversarial examples is to block their transferability. The question is: can we train the classifier to discard adversarial examples, while maintaining the accuracy on the clean data?} We study this problem in the black-box setting and observed that adversarial examples transferred from the substitute classifier can be viewed as the noisy version of the adversarial examples generated on the target classifier. In common ML classifiers, the predictions vary smoothly around the input samples, i.e., they classify the noisy samples into the same label as the clean samples. This property enables the transferability of the adversarial examples. To block transferability, we propose a training method such that, as the input is more perturbed, the classifier shows lower confidence on the original label and instead declares that the input is ``invalid''. During the training phase, the classifier examines the distinction between the clean and adversarial data, and learns to discard the adversarial examples, while keeping the clean inputs.

\vspace{0.1cm}
In this paper, we make the following contributions: 
\begin{itemize}
	\item It is reasonable to assume that the adversary may not have any access to the target system. Therefore, considering the black-box model, we propose a more restricted adversary's model called {\it Blind Model}. In the blind model, the adversary only possesses a small dataset and, unlike the black-box model, cannot query the target classifier. Using the transferability property of adversarial examples, we develop attacks on learning systems under both black-box and blind models.
	
	\item We experimentally evaluate the robustness of the black-box learning systems against the adversarial examples. In particular, we consider 1) online learning systems from Amazon and Microsoft oracles, 2) deep neural networks, and 3) neural networks trained to be robust~\cite{goodfellow2014explaining}. We show that even under the blind model, the adversary can mount both the misclassification and targeted attacks. This further represents a more serious attack vector on black-box learning systems. 
	
	\item We then propose a defense framework for black-box systems to block the transferability of the adversarial examples. Essentially, we augment the output class set of the classifier with an additional label called \nl. During the training phase, we generate adversarial examples and train the ML algorithm to label them as \nl with appropriate probabilities according to the amount of perturbation. We show that the proposed method can effectively reject adversarial examples, by classifying them as \nl, while classifying the clean data as their original labels.
\end{itemize}

As an illustration, Fig. \ref{fig:images} compares the approach of \nl labeling with the original labeling. The ML classifier assigns a probability vector to each image. As the input is more perturbed, the classifier's belief in the ground-truth label decreases and eventually, with enough perturbation, the adversary can cause the classifier to misclassify the input with high confidence. However, with \nl labeling, as the input is more perturbed, the classifier assigns higher probability to the \nl label, and thus discards the adversarial example.

\section{Preliminaries}\label{sec:prel}

Traditional ML models can be partitioned into two categories of supervised or unsupervised learning, which are the tasks of inferring a function from labeled training data or inferring a function to describe hidden structure from unlabeled data, respectively \cite{mohri2012foundations}. Similar to previous works on adversarial learning \cite{szegedy2013intriguing,goodfellow2014explaining,carlini2016towards,papernot2016limitations,papernot2016practical,papernot2016transferability}, we restrict ourselves to the problem of learning multi-class classifiers in supervised settings.


A classifier is a function that maps the input feature vector $X$ to the output probability vector $Z$, where $Z_i$, the $i$-th element of the vector $Z$, represents the probability that the input belongs to the $i$-th class. For each input, the output label is the class with the highest probability. The training data is composed of the labeled samples $(X,y)$. We assume that input features are continuous and bounded and, without loss of generality, assume they are bounded between $0$ and $1$, i.e., $X_i \in [0,1]$. 

\begin{figure}[h]
	\centering
	\includegraphics[width=0.33\textwidth]{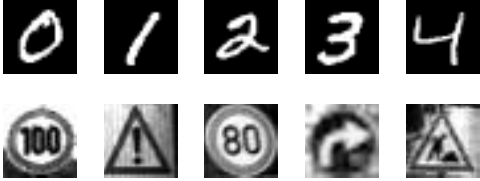}
	\caption{Representative samples from MNIST dataset (top) and GTSRB dataset (bottom).}\vspace{-0.15cm}
	\label{fig:MNIST_GTSRB}
\end{figure}

Throughout this paper, we demonstrate the results on two ML datasets, MNIST~\cite{lecun1998mnist} and German Traffic Signs Recognition Benchmark (GTSRB)~\cite{stallkamp2012man}.
MNIST is a dataset of handwritten digits with $70,000$ gray-scale images of size $28\times28$, split into three groups of $50,000$ training samples, $10,000$ validation samples, and $10,000$ test samples. The task involves the classification of images of digits into one of $10$ classes, corresponding to digits $0$ to $9$. 
GTSRB is a dataset of color images of 43 traffic signs, with $39,209$ training samples, and $12,630$ test samples. Images vary in size; we resize all images to the same size as MNIST inputs ($28\times28$ gray-scale images) and pass them through a histogram equalization filter. We also randomize the ordering of the training data and use the first $5,000$ samples as the validation set. 
Some representative samples of each dataset are shown in Figure \ref{fig:MNIST_GTSRB}.
We present the experimental results on Deep Neural Networks (DNN). The network structures used for MNIST and GTSRB datasets are given in Appendix~\ref{apx:DNNs}.

\section{Adversarial Examples}

In this section, we provide the definition of adversarial examples and review different methods for generating adversarial examples and defending against them. We describe the techniques which are most relevant to our context and will explore other techniques in more details in Section \ref{related}.

\subsection{Definition of Adversarial Examples}

ML classifiers are vulnerable to adversarial perturbations, i.e., an adversary can slightly perturb an input such that the classifier will output the wrong label, yet a human will classify it correctly~\cite{szegedy2013intriguing}. Such modified inputs are called {\it adversarial examples}. As ML systems are being increasingly integrated into critical and security-sensitive applications, the existence of such adversarial examples poses a major security threat to these systems. 
The feasibility of generating adversarial examples have been demonstrated in real-world applications \cite{papernot2016practical,papernot2016transferability,hosseini2017deceiving}. Also, in \cite{kurakin2016adversarial}, it has been shown that generating such inputs is practical in physical world scenarios, where the samples, for instance, are perceived through a camera.


\subsection{Generating Adversarial Examples} \label{sec:prior_attack}

Attacks based on adversarial examples have one of the following goals~\cite{papernot2016limitations}: 1) Misclassification Attack: The adversary causes the classifier to output a label different from the original label, and 2) Targeted Attack: The adversary causes the classifier to output a specific label. For generating an adversarial example, the adversary perturbs the input so as to decrease the confidence of the classifier on the original label (for misclassification attack) or increase the confidence on the target label (for targeted attack). The perturbation however should be small, in order to remain unnoticeable by human observer.
Formally, the adversary's problem is: given  $(X,y)$, find $X^* = X + \delta X$ such that the classifier maps $X^*$ into the adversary-desired label and $\|\delta X\| \leq \epsilon \cdot \delta_{\max}$, where $\delta_{\max}$ denotes the maximum possible perturbation. When $\|\delta X\| = \epsilon \cdot \delta_{\max}$, we say $X^*$ is an $\epsilon$-perturbation of $X$.

\begin{figure}[t]
	\centering
	\includegraphics[width=0.25\textwidth]{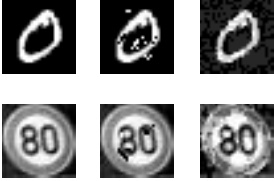}
	\caption{Adversarial examples on a DNN. Examples are generated on one sample of the MNIST dataset (top) and GTSRB dataset (bottom). From left to right: original image, adversarial examples generated using $\mathtt{Grad_0}$ and FGS methods.}\vspace{-0.3cm}
	\label{fig:adv_compare}
\end{figure}

In~\cite{goodfellow2014explaining}, Goodfellow et al. proposed a gradient-based method for the misclassification attack called Fast Gradient Sign (FGS) method. Given the classifier, FGS method generates adversarial examples with respect to the $L_{\infty}$ norm constraint. In this paper, we adopt the technique for the $L_0$ norm constraint, i.e., we generate adversarial examples with changing few input features, called {\it adversarial features}. We call our method as $\mathtt{Grad_0}$. 
We also propose a Greedy Method for the targeted attack, which follows the gradient-based approach, but selects one adversarial feature at a time. In Appendix~\ref{apx:Adv_gen}, we will explain different methods for generating adversarial examples in detail. Figure~\ref{fig:adv_compare} represents some adversarial examples on a DNN. The examples are generated using the gradient-based method with different norm constraints on one sample of the MNIST and GTSRB datasets. 

\subsection{Robust Training Method}
One natural countermeasure against the adversarial examples is to change the training procedure to reflect the fact that perturbations of each training sample should be classified as the original class. This method is alternatively called as adversarial training~\cite{goodfellow2014explaining} or robust training~\cite{shaham2015understanding}. Mathematically, this is equivalent to training with the objective function:\vspace{-0.05cm}
$$\theta^*=\argmin_{\theta} \alpha \ell(X,y;\theta) + (1-\alpha) \ell(X^*,y;\theta),$$
where $\ell(X, y; \theta)$ denotes the loss of the classifier with parameters $\theta$ on $(X, y)$, $X^*$ is an adversarial example generated using a gradient-based method and the parameter $\alpha\in[0,1]$ balances the training on the clean and adversarial samples. This approach iteratively creates a supply of adversarial examples and trains the classifier to correctly classify them. It has been shown that the modified objective function makes the classifier more resistant to adversarial examples. 
For consistency, we call the robust training as $\mathtt{Robust_0}$ or $\mathtt{Robust_{\infty}}$ methods, according to the type of the norm constraint that is used for generating the adversarial examples.


%


\section{The Proposed Attacks}\label{Attacks}

In this section, we present the attack methods and provide the experimental results on attacking various ML classifiers. 

\subsection{Threat Models} \label{sec:adv_model}

We present the results under two threat models which differ based on the adversary's capability of querying the target system. The models are described in the following.
\begin{itemize}
	\item {\bf Black-Box Model:} The adversary possesses a limited set of labeled data, with the same distribution as the target's training data, and also has an oracle access to the system, i.e., she can query the classifier for any input and get the corresponding label.
	
	\item {\bf Blind Model:} The adversary possesses a limited set of labeled data, with the same distribution as the target's training data. However, unlike the black-box model, the adversary is blind to the target system, i.e., she does not have any kind of access to the classifier.
\end{itemize}


In black-box model, the distinction between knowledge of labels and the output probability vectors is important, since, in real world applications, users often have access to output labels than output probability vectors; the adversary is however weaker because labels contain much less information about the classifier's behavior~\cite{papernot2016practical}.
%
%


\subsection{Training Substitute Classifier}\label{gen_attack}

\begin{figure}[t]
	\centering
	\includegraphics[width=.48\textwidth]{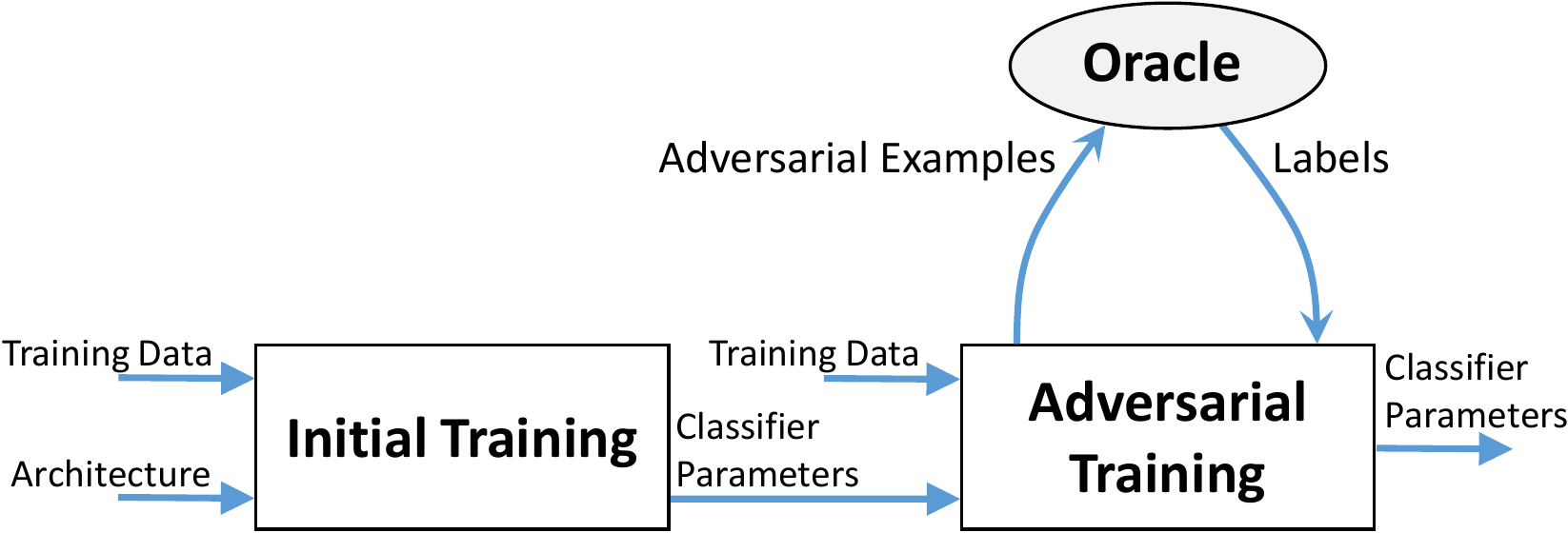}
	\caption{Block diagram of the proposed method for training the substitute classifier in the black-box model. In the adversarial training phase, we alternatively train the classifier on the clean data and generated adversarial examples, for which we get the labels by querying the oracle.}
	\label{fig:BlackBox}
\end{figure}

In the following, we present the training method of the substitute classifier under the black-box and blind models. 

\vspace{0.2cm}
\noindent{\it Black-Box Model.} 
We intend to train the substitute classifier in a way that it can mimic the decision boundaries of the target classifier. For this, we iteratively update the classifier and query the oracle for labels of the new adversarial examples generated based on the current classifier parameters. 
However, since the adversary possesses only a small set of labeled data, during the first few epochs, the classifier accuracy is very low. As a result, the generated adversarial examples do not lie on the decision boundaries of the target classifier. Therefore, to generate higher quality adversarial examples, we need to first train the substitute classifier on the clean data for a certain number of epochs, which we refer to as initial training.

After the initial training, if we only train the classifier on the generated adversarial examples, the classifier starts to perform poorly on the clean data. 
Therefore, we propose to alternatively train the substitute classifier on the clean and adversarial examples to keep the classifier fairly accurate on the clean data and also enable it to learn the decision boundaries of the target classifier. Figure~\ref{fig:BlackBox} depicts the block diagram of the proposed method for training the substitute classifier in the black-box model.

\noindent{\it Blind Model.} 
The procedure is similar to the black-box model, except that, since the adversary cannot query the target classifier, she only does the initial training for the substitute classifier. 

\vspace{0.2cm}
In both the black-box and blind models, the adversary generates adversarial examples on the substitute classifier and transfers them to the target classifier. For generating adversarial examples, a gradient-based method is used for misclassification, and the greedy method is used for targeted attack. If the adversary can successfully attack the substitute classifier, the generated samples are used as the adversarial examples for the target classifier. 


\subsection{Experimental Results}\label{sec:attack_current}

We present the results for the blind model and for adversarial examples generated by $\mathtt{Grad_0}$ method. The attacks under the black-box model are strictly more successful than those in the blind model. Also, the adversarial examples generated by the FGS method perform similar to the $\mathtt{Grad_0}$ method. We train the substitute classifier for $50$ epochs, and then generate adversarial examples on available data with a perturbation ranging from $2.5\%$ to $20\%$ of the maximum possible perturbation. 
In experiments with the MNIST dataset, we assume that the adversary possesses $100$ labeled samples for training the substitute classifier and an additional $50$ labeled samples for validation. For GTSRB, we assume that the adversary possesses $250$ labeled samples, of which $150$ samples are used for training and $100$ labeled samples for validation. For training the substitute classifier, we use the data from MNIST and GTSRB validation set. The number of samples is chosen such that we will have sufficient data from all classes. Using more data samples increases the adversary's success rate. 

We apply the attacks on a DNN and also on robust DNNs trained with different types of adversarial examples. For training the robust classifiers, we set $\alpha=0.5$ and use $0.1$- and $0.3$-perturbed adversarial examples, respectively for MNIST and GTSRB datasets. The classifiers are trained on the DNN structures described in Appendix~\ref{apx:DNNs}. We perform the attacks also on two ML-as-a-service platforms: 1) Amazon Web Services Oracle: We train a classifier using the Amazon ML platform which is part of the Amazon Web Services (AWS)\footnote{\url{https://aws.amazon.com/machine-learning/}}, and 2) Microsoft Azure Oracle: We set up a neural network classifier on the Microsoft Azure ML platform\footnote {\url{https://studio.azureml.net/}}. 

\begin{table}[t]
	\centering
	\caption{Test accuracies of different classifiers trained on MNIST and GTSRB datasets.}
	\begin{tabular}{ |l|c|c| }
		\hline
		ML Classifiers & MNIST & GTSRB \\
		\hline
		\hline
		DNN & $99.35\%$ & $97.77\%$ \\
		\hline
		DNN trained with $\mathtt{Robust_0}$ method & $98.81\%$ & $97.05\%$ \\
		\hline
		DNN trained with $\mathtt{Robust_{\infty}}$ method & $99.39\%$ & $96.80\%$ \\
		\hline
		Amazon Oracle & $92.00\%$ & $72.81\%$ \\
		\hline
		Microsoft Oracle & $97.73\%$ & $85.76\%$ \\
		\hline
	\end{tabular}
	\label{table:Accs}
\end{table}

Table~\ref{table:Accs} provides the test accuracies of the different classifiers trained on MNIST and GTSRB datasets. It can be seen that $\mathtt{Robust_{\infty}}$ method slightly improves the test accuracy on MNIST, but decreases the accuracy on GTSRB. The $\mathtt{Robust_0}$ method however decreases the accuracy on both datasets. The classifiers trained on Amazon platform have relatively very low accuracy, while the classifiers trained on Microsoft platform yield moderate accuracy on MNIST and low accuracy on GTSRB dataset.

\begin{figure}[t]
	\centering
	\includegraphics[width=.45\textwidth]{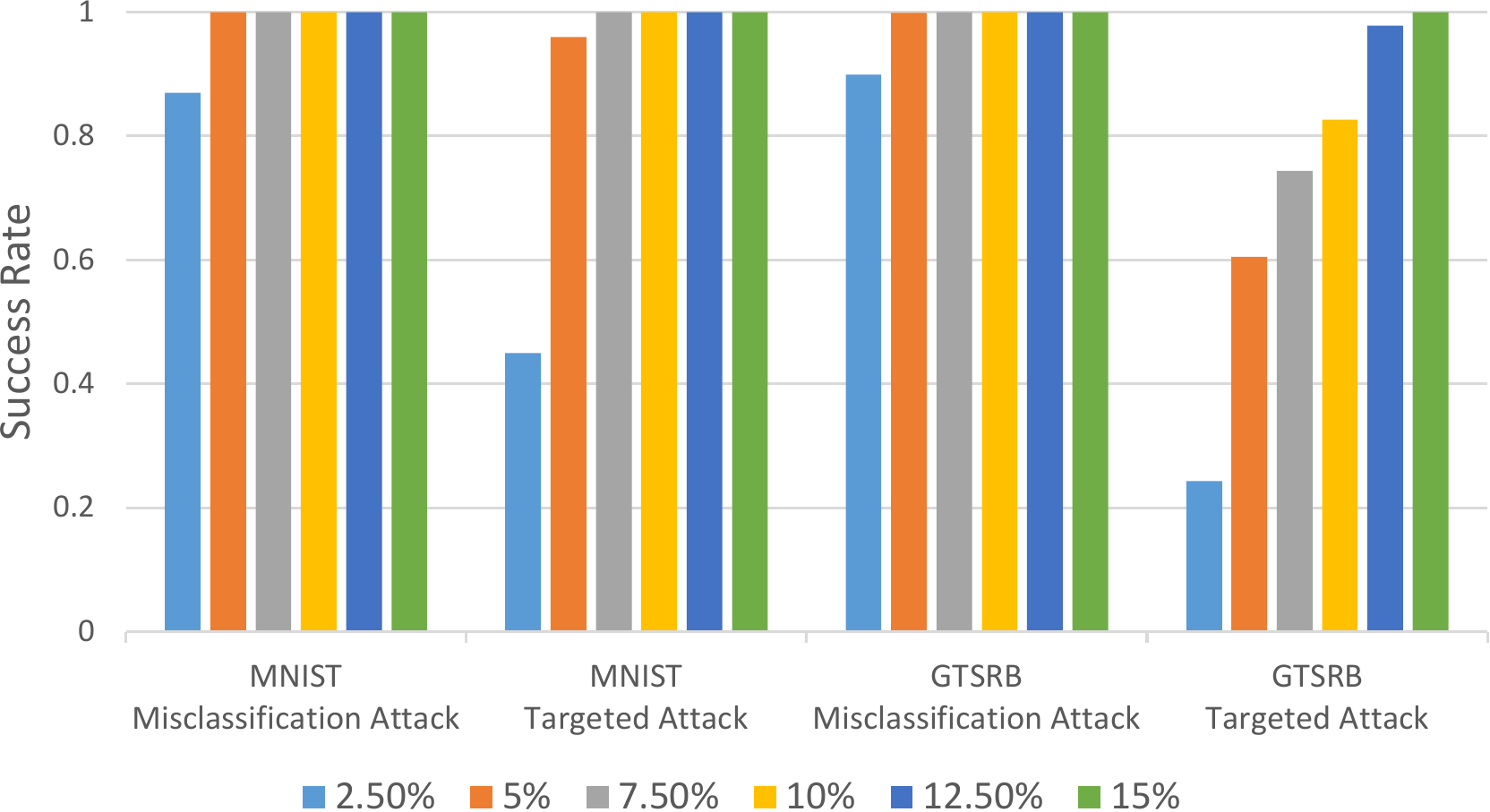}
	\caption{Adversary's success rate in generating adversarial examples on substitute classifier for different amount of perturbations. Note that since the adversary has full knowledge about the substitute classifier, we expect to have high adversary's success rate.}\vspace{-0.1cm}
	\label{fig:blindSuccess}
\end{figure}

In the following, we show that we can generate adversarial examples on the substitute classifier and also transfer them to the target classifier. Figure \ref{fig:blindSuccess} shows the adversary's success rate in generating the adversarial examples on the substitute classifier. It can be seen that by allowing higher perturbation ($\epsilon$), the adversary's success rate quickly increases. Recall that the adversarial example $X^*=X + \delta X$ is an $\epsilon$-perturbation of $X$, if $\|\delta X\| = \epsilon \cdot \delta_{\max}$, where $\delta_{\max}$ is the maximum possible perturbation. 

\begin{figure*}[t]
	\begin{subfigure}{.5\textwidth}
		\centering
		\includegraphics[width=0.85\linewidth]{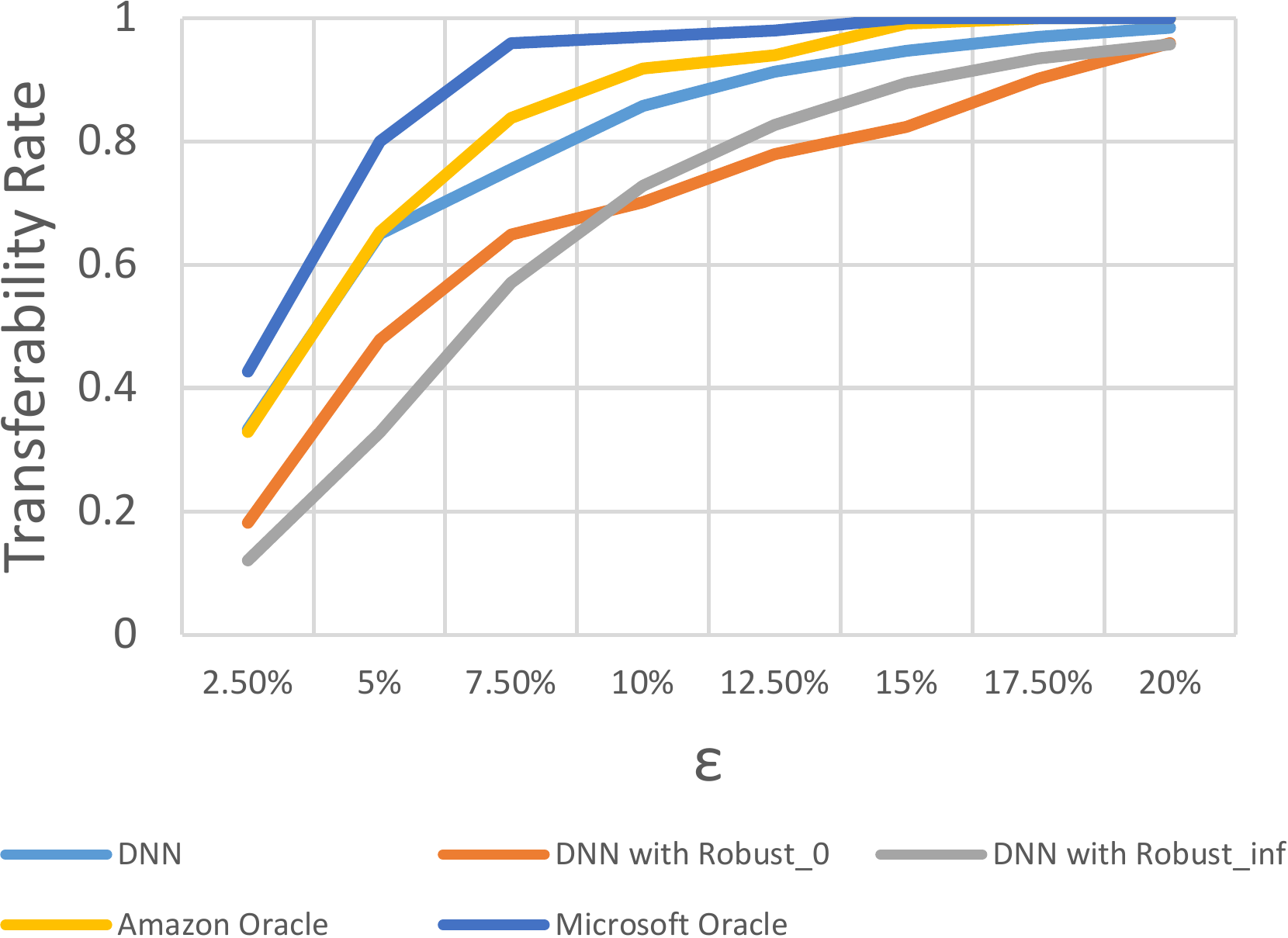}\vspace{0.1cm}
		\caption{Misclassification attack on classifiers trained on MNIST dataset.}
	\end{subfigure}
	\begin{subfigure}{.5\textwidth}
		\centering
		\includegraphics[width=0.85\linewidth]{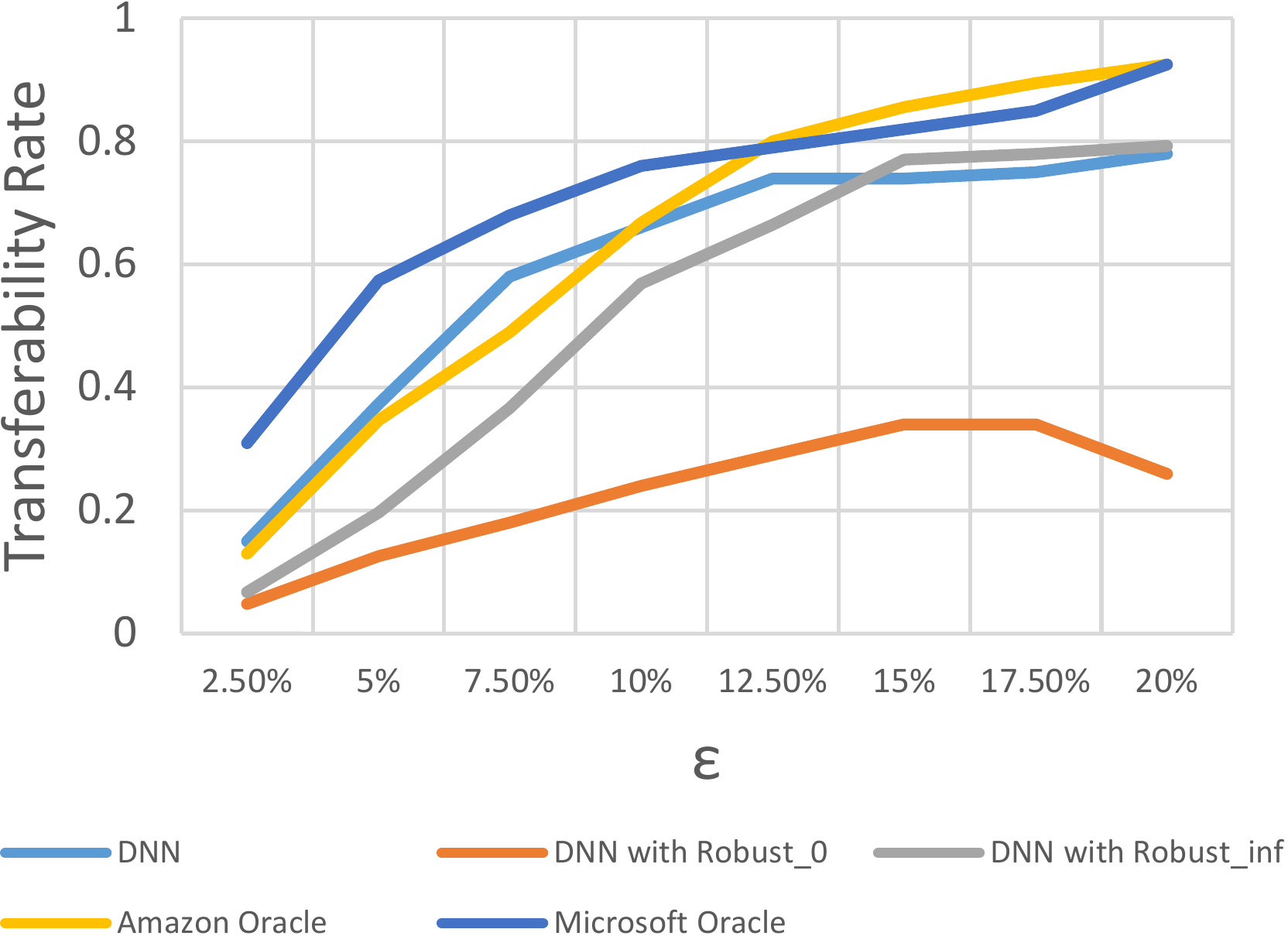}\vspace{0.1cm}
		\caption{Targeted attack on classifiers trained on MNIST dataset.}
	\end{subfigure}\vspace{0.4cm}
	\begin{subfigure}{.5\textwidth}
		\centering
		\includegraphics[width=0.85\linewidth]{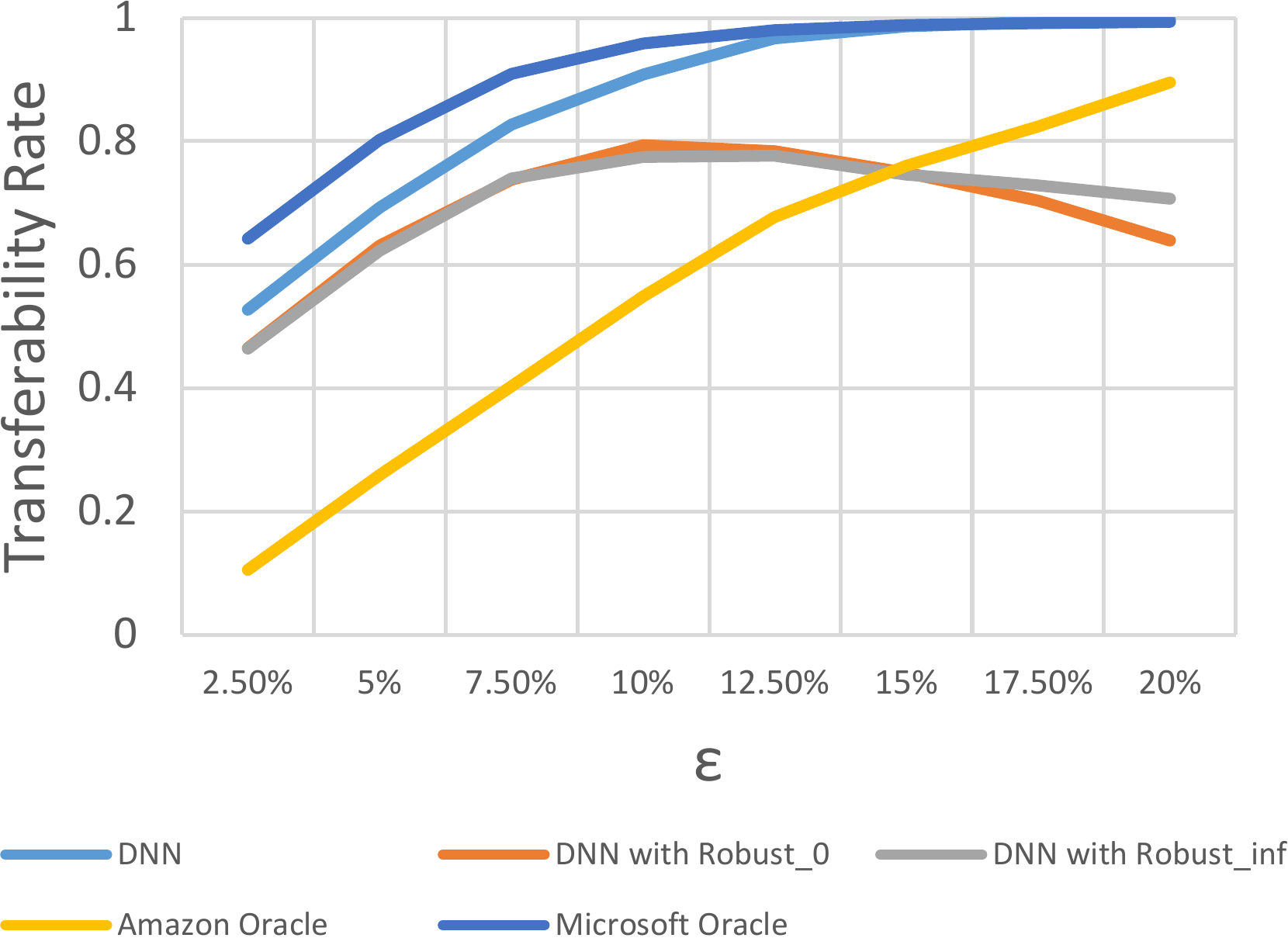}\vspace{0.1cm}
		\caption{Misclassification attack on classifiers trained on GTSRB dataset.}
	\end{subfigure}
	\begin{subfigure}{.5\textwidth}
		\centering
		\includegraphics[width=0.85\linewidth]{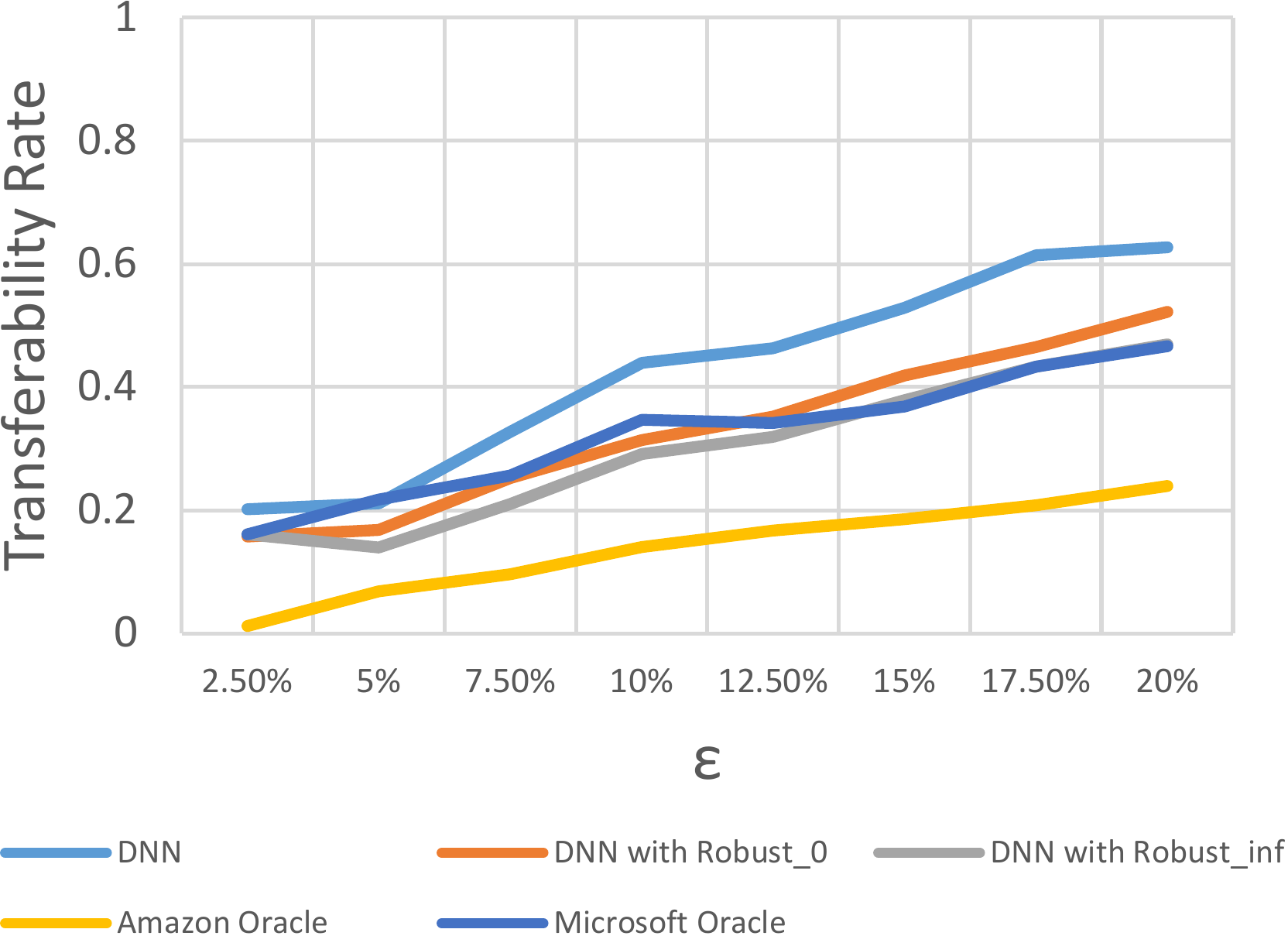}\vspace{0.1cm}
		\caption{Targeted attack on classifiers trained on GTSRB dataset.}
	\end{subfigure}
	\caption{Transferability rate of adversarial examples to different ML classifiers under the blind model.}
	\label{fig:attack_others}
\end{figure*}

Figure \ref{fig:attack_others} shows the transferability rate of the adversarial examples to different classifiers for misclassification and targeted attacks. The transferability rates are higher in general for MNIST and, as expected, misclassification attack is more successful than the targeted attack. For GTSRB dataset, the transferability rate is relatively low for Amazon-based classifier, especially for the targeted attack. The reason might be that since the Amazon classifier performs poorly on clean data, it is also difficult to force it to output a specific target label for an adversarial example. 

The results show that robust training provides some resistance against the adversarial examples under the blind model. The transferability rates however increase in the black-box model, where the adversary is allowed to query the target classifier. Recall that in blind model, the adversary only possesses a few labeled samples, yet we show that we can achieve high transferability rates. Our results are significantly better than the current state of the art for the targeted attack. For example, in~\cite{papernot2016practical}, for the MNIST dataset and at $\epsilon=28.57\%$, the average transferability is $55.53\%$, while we achieved $66\%$ and $78\%$ transferability rates  on the DNN classifier for $\epsilon=10\%$ and $20\%$, respectively, without querying the target classifier.

\section{The Proposed Defense Method}

In this section, we propose the \nl labeling method for defending against adversarial examples under black-box settings and evaluate its robustness through the simulations. 

\subsection{Our Goal and Methodology}

The attacks based on adversarial examples on black-box learning systems rely on the transferability property. Therefore, the key to protect ML oracles is to block the transferability of adversarial examples. Since the decision boundaries of two classifiers trained on equally distributed datasets are approximately the same, an adversarial example for one classifier can be viewed as the noisy version of the adversarial examples for another classifier. In common ML classifiers, the predictions vary smoothly around the input samples, i.e., they classify the noisy samples into the same label as the clean samples. This enables the transferability of the adversarial examples. To block transferability, we propose to train the classifier such that, as the input noise level increases, the classifier shows lower confidence on the original label and instead declares the input as invalid. The method should however maintain the accuracy on the clean inputs. 

To this end, similar to previous methods~\cite{goodfellow2014explaining,lyu2015unified,shaham2015understanding}, we  train the classifier also on adversarial examples generated from the clean data. However, since our goal is to identify and discard the malicious data rather than predicting their original label, we train the classifier to label the adversarial examples as ``not-a-valid-input". But, the distribution of the test data is also likely to slightly deviate from that of the training data. Therefore, if we train the classifier to discard all perturbed inputs, regardless of the amount of perturbation, the classifier will perform poorly on clean test data. Hence, we need a measure to decide how likely an input sample is adversarial. 
To this end, as illustrated in Figure~\ref{fig:images}, we augment the output class set with a \nl label and train the classifier such that the \nl probability assigned to each input represents the belief of classifier that it is adversarial. 


\subsection{Techniques Used in Proposed Method}

We use the following techniques in the proposed training method.

\vspace{0.2cm}
\noindent {\bf Label Smoothing.}
We use the label-smoothing regularization technique proposed in~\cite{szegedy2016rethinking}. In this technique, we train the classifier on modified output probability vectors, for which the probability of the ground-truth label is decreased and the rest of the probability is distribute uniformly to other labels. Assume that we have $K$ classes. If we decrease the probability of the ground-truth label to $q$, each one of other labels will get $\frac{1-q}{K-1}$ probability. This technique encourages the classifier to be less confident on clean data and is shown to help the generalization. 

\vspace{0.2cm}
\noindent {\it {\bf MG Method:} Misclassification Attack Using Greedy Method.} 
We use MG method to generate adversarial examples for the misclassification attack. We adopt the gradient-based method, but take a greedy approach, i.e., we select and modify one feature at a time until the input is successfully misclassified.

\vspace{0.2cm}
\noindent {\it {\bf STG Method:} Smooth Targeted Attack Based on Gradient Method.} 
We train the classifier	on the adversarial examples generated for the targeted attack. However, unlike current methods, we set a soft target, i.e., the target output probability vector is $\frac{1}{K}$ for the $K$ valid classes. We use the $L_0$ norm constraint for generating adversarial examples. The details of MG and STG methods are provided in Appendix~\ref{apx:Adv_gen}.

\subsection{The NULL Labeling Method}

\begin{figure}[t]
	\centering
	\includegraphics[width=0.475\textwidth]{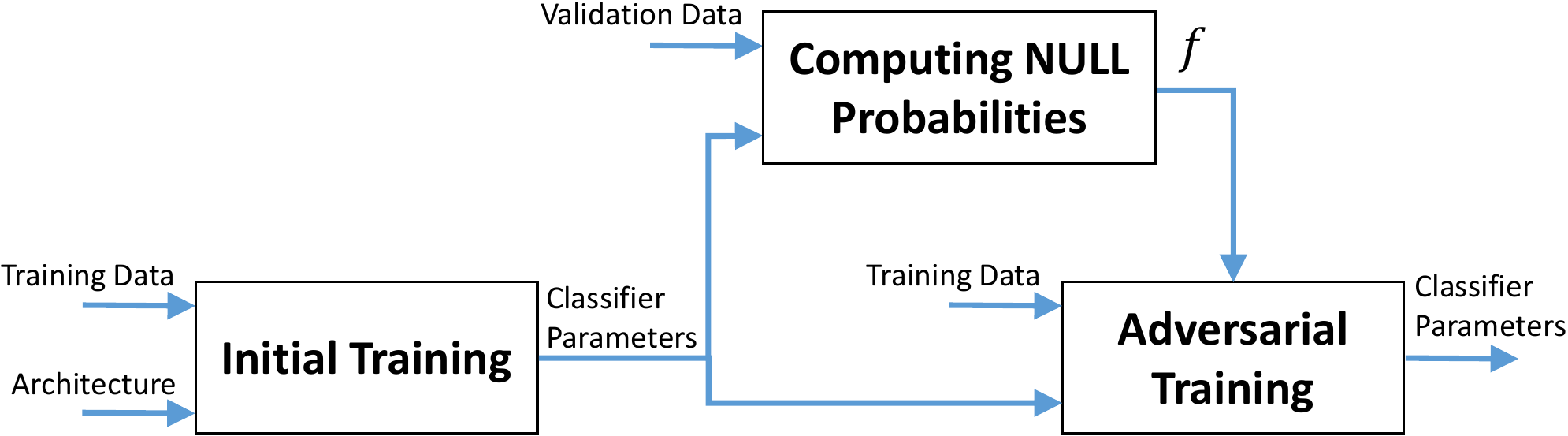}
	\caption{Block diagram of \nl labeling method. In adversarial training phase, the classifier is trained on both clean and adversarial data. The classifier learns to map adversarial examples to the \nl label with a probability according to their amount of perturbation.}\vspace{-0.4cm}
	\label{fig:null_method}
\end{figure}

The proposed method is composed of three steps: 1) initial training, 2) computing \nl probabilities for adversarial examples generated with different amount of perturbations, and 3) adversarial training. 
Figure \ref{fig:null_method} depicts the block diagram of the proposed training method. In the following, we describe our method in details.

\vspace{0.2cm}
\noindent {\bf Initial Training.}
Since the classifier parameters are initialized randomly, the adversarial examples generated at the beginning do not resemble those of the final classifier. Therefore, we initially train the classifier only on the clean data to generate higher quality adversarial examples.

\vspace{0.1cm}
\noindent {\bf Computing \nl Probabilities for Adversarial Examples.}
We then need to compute a function $f$ that assigns a proper \nl probability $p_{\nl}$ to the adversarial examples, generated during the training phase, according to their amount of perturbation. For computing the function $f$, we apply the MG method on validation data. Note that MG method takes a greedy approach for generating adversarial examples for the misclassification attack. Therefore, we succeed by perturbing relatively few number of input features.

\begin{figure}[t]
	\centering
	\includegraphics[width=0.75\linewidth]{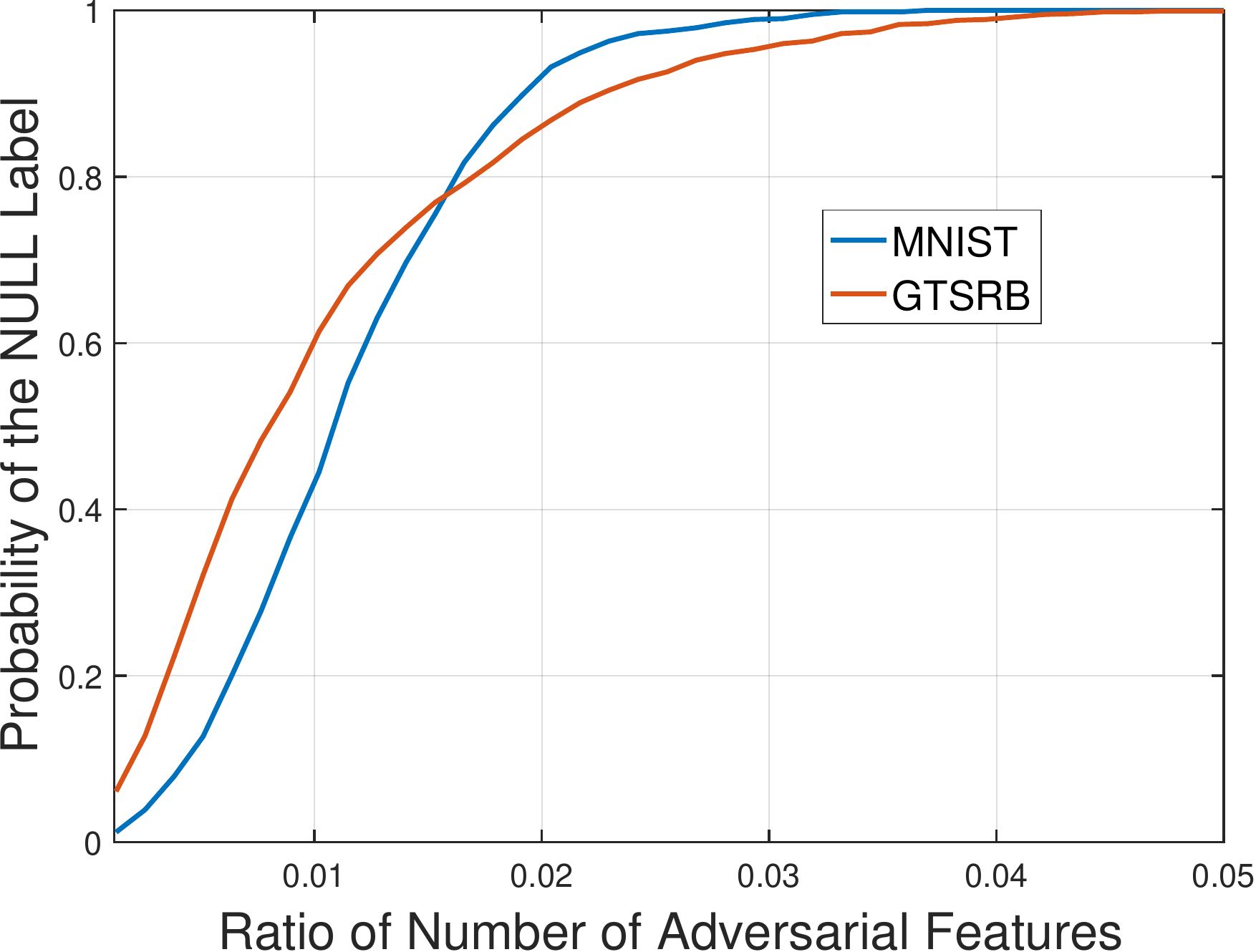}
	\caption{Probability of the \nl label for an adversarial example versus $\epsilon$, the ratio of the number of adversarial features, for DNNs trained on MNIST and GTSRB datasets.}\vspace{0.3cm}
	\label{fig:null_plot}
\end{figure}

If a fraction $\alpha$ of the generated adversarial examples have perturbation less than $\epsilon$, we set the \nl probability for $\epsilon$-perturbed inputs to be $\alpha$. This shows with what probability we can cause the classifier to misclassify an input, if we are allowed to perturb $\epsilon \cdot |X|$ number of features. 
We compute the function $f$ on the classifier obtained after the initial phase and use it for the entire adversarial training phase.
Figure \ref{fig:null_plot} depicts the function $f$ for DNNs trained on MNIST and GTSRB datasets.

\vspace{0.1cm}
\noindent {\bf Adversarial Training.}
During the adversarial training phase, in each epoch and for each sample, we train on the clean sample or its corresponding adversarial example, with probabilities $\alpha$ and $1-\alpha$, respectively. 
Adversarial examples are generated using the STG method, with a number of adversarial features, which is chosen uniformly at random between one and $N_{\max}$, where $N_{\max}$ is the minimum number for which $f(\frac{N_{\max}}{|X|})=1$. The STG method modifies inputs such that the output probability increases for all valid labels other than the ground-truth label. The classifier is however trained to map the adversarial examples to a vector which shows high probability at the \nl label. Essentially, the classifier adaptively finds the problematic directions around each input sample and corrects itself by nullifying them. 

\begin{figure}[h]
	\centering
	\includegraphics[width=0.47\textwidth]{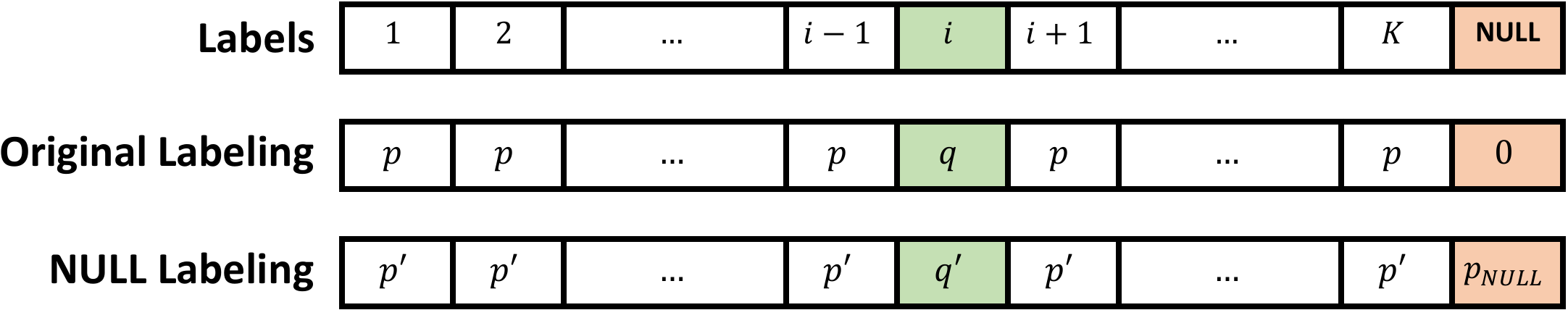}
	\caption{Illustration of the \nl labeling with label smoothing technique. Assume that there are $K$ valid classes and the $i$-th label is the ground-truth. Label smoothing technique assigns a probability $q$ to the ground-truth label and distributes the rest of the probability uniformly to other valid labels, i.e., $p=\frac{1-q}{K-1}$. Similarly, for an adversarial example with \nl probability $p_{\nl}$, we set $q'=q(1-p_{\nl})$ and $p'=\frac{(1-q)(1-p_{\nl})}{(K-1)}$. Note that in Figure~\ref{fig:images}, for simplicity, we did not incorporate the label smoothing technique in illustration of the \nl labeling method.}
	\label{fig:null_ex}
\end{figure}

Formally, we train the network by repeatedly solving the following optimization problems:
$$
\Resize{8.5cm}{
\begin{cases}
{\delta X}^*=\argmin_{\delta X} \ell(X+\delta X,Z_T;\theta), \;\; \mbox{s.t. }\|\delta X\|_0 \sim U[1,N_{\max}]\\
\theta^* = \argmin_{\theta} \alpha \ell(X,Z;\theta) + (1-\alpha) \ell(X+{\delta X}^*,Z_A;\theta)
\end{cases}}$$
where $U$ denotes the uniform distribution and $\alpha$ represents how often we train on original training data and controls the trade-off between test accuracy and the ability to detect adversarial examples. 
The vector $Z_T$ is the smooth target probability vector that we use to generate adversarial examples, which is zero at the \nl label and $\frac{1}{K}$ otherwise. The vector $Z_A$ is however the desired output probability vector and is obtained as illustrated in Figure~\ref{fig:null_ex}. We first compute the \nl probability as $p_{\nl}=f(\frac{\|{\delta X}^*\|_0}{|X|})$ and modify the probability of the ground-truth label from $q$ to $q'=q(1-p_{\nl})$. The probability of the rest of the labels will then be $\frac{(1-q)(1-p_{\nl})}{(K-1)}$.

\subsection{Experimental Results} \label{simulations}

\begin{figure*}[t]
	\centering
	\begin{subfigure}{.45\textwidth}
		\centering
		\includegraphics[width=0.75\linewidth]{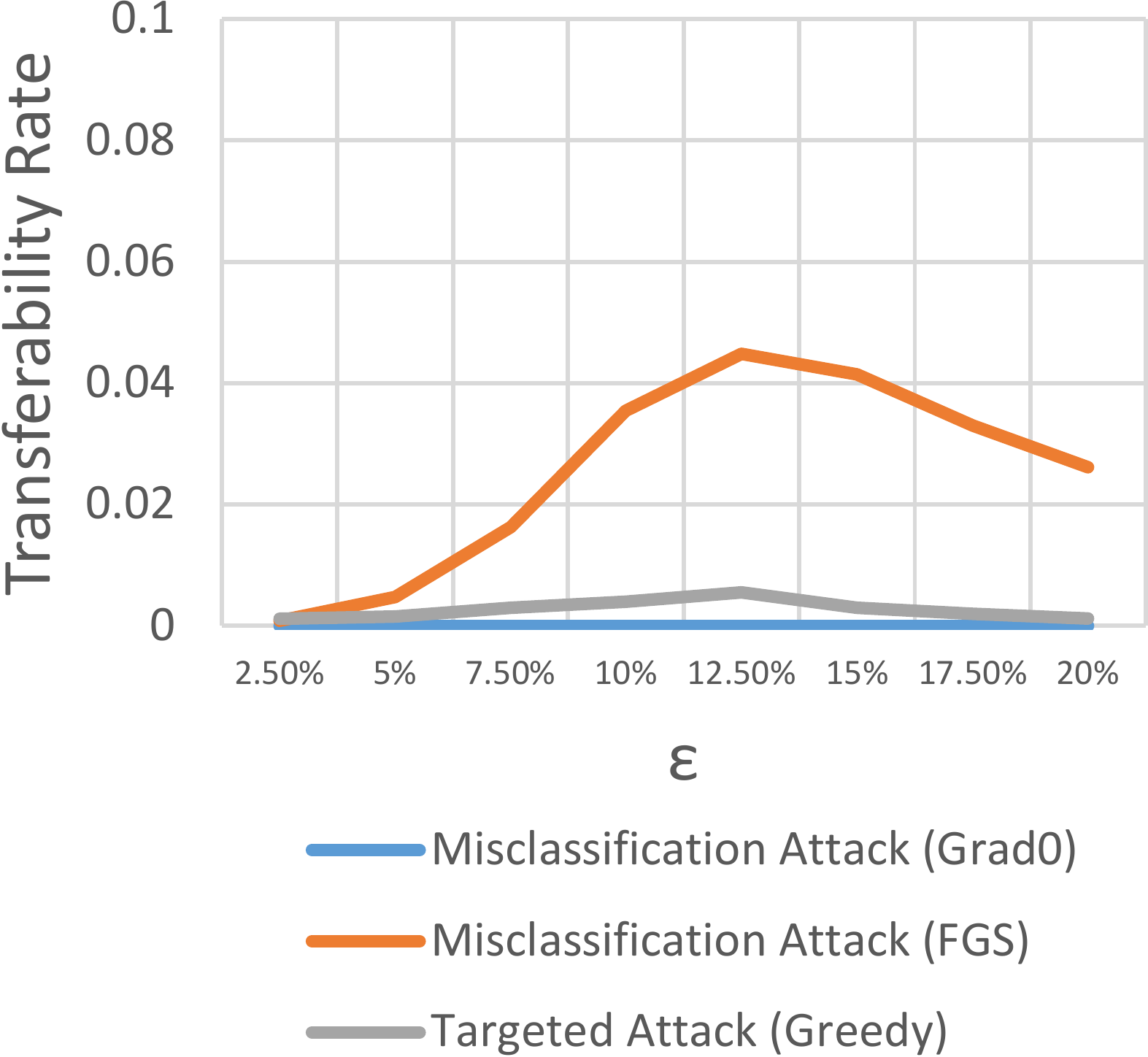}\vspace{0.075cm}
		\caption{Transferability rates under the blind model.}
	\end{subfigure}\;\;\;\;\;
	\begin{subfigure}{.45\textwidth}
		\centering
		\includegraphics[width=0.75\linewidth]{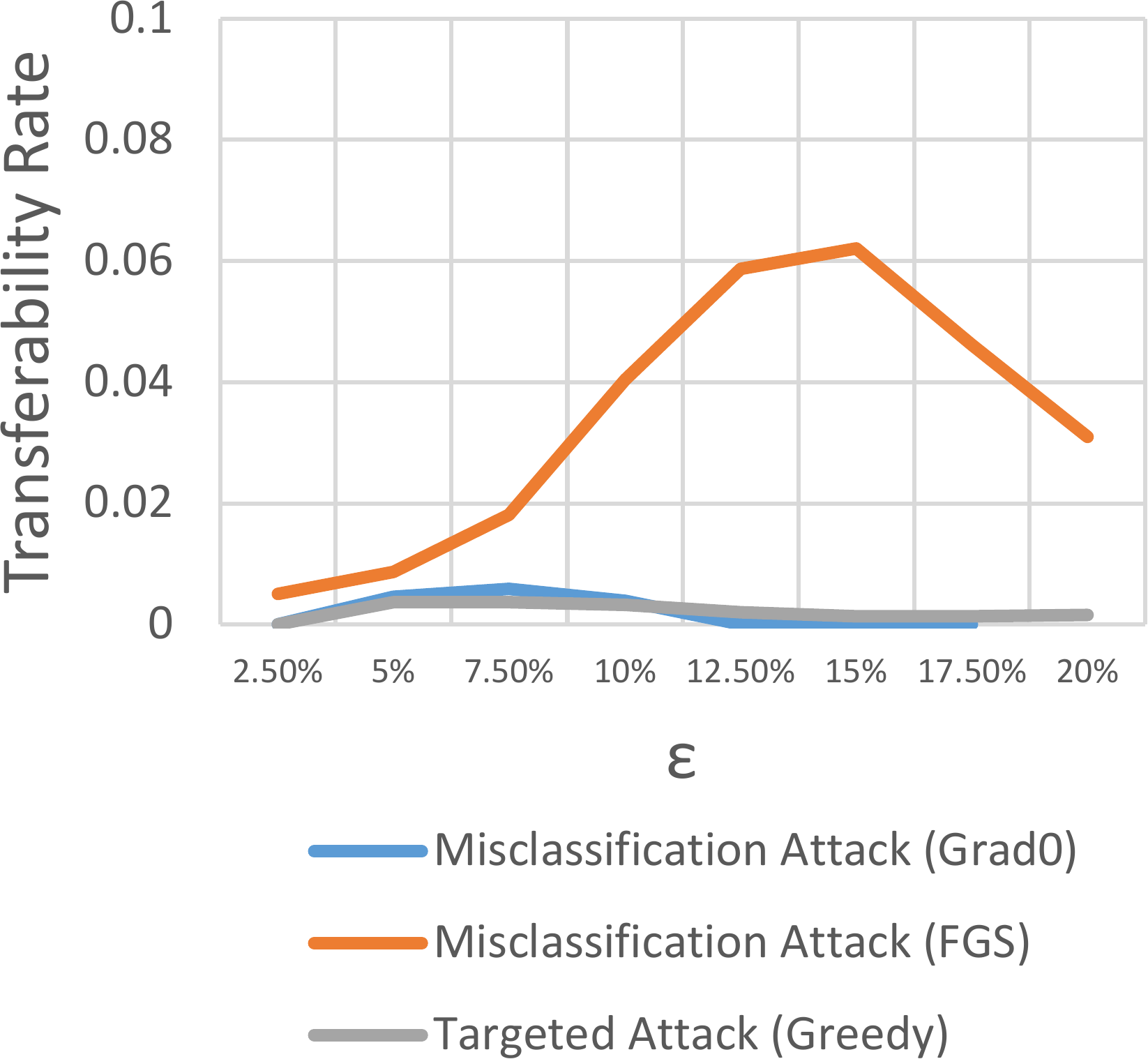}\vspace{0.075cm}
		\caption{Transferability rates under the black-box model.}
	\end{subfigure}
	\caption{Transferability rate of adversarial examples to the classifier trained with the \nl labeling method on GTSRB dataset.}
	\label{fig:attack_null}
\end{figure*}

Note that in attacks based on adversarial examples, the adversary seeks to avoid being detected. Therefore, {\it an adversarial example which is misclassified to the \nl label is considered as a failed attack}, because a \nl labeled input signals a suspicious activity. Specifically, a misclassification attack is successful if the classifier maps the input to a label, other than the original or the \nl labels. Similarly, in targeted attack, the \nl label cannot be considered as a possible target label. 

We train the classifier for only $1$ epoch in initial training and $200$ epochs in adversarial training phase. Fixing $\alpha=0.5$, the only hyperparameter introduced by our method is the label smoothing parameter $q$, which is inherited from the label smoothing technique. We set $q$ as the value that maximizes the validation accuracy, and obtained $q=0.9$ for MNIST dataset and $q=0.6$ for GTSRB dataset. We output the classifier parameters that resulted in the highest validation accuracy. 
The test accuracies for MNIST and GTSRB datasets are $99.46\%$ and $97.37\%$, respectively. Compared to a DNN with the same structure, our proposed method slightly increases the test accuracy for MNIST, and slightly decreases the test accuracy for GTSRB. Also, for both datasets, our method outperforms the adversarial training methods. 

The attack settings are the same as those in Section~\ref{sec:attack_current}. 
In the black-box model, we train the substitute classifier for $50$ epochs in initial training and $100$ epochs in adversarial training. 
During the adversarial training phase, the adversary queries the oracle with generated adversarial examples and, for some of them, receives the \nl label. We take two approaches for training the substitute classifier. The first approach is to discard those samples, and the second approach is to augment the output class set and train the classifier to map those examples into the \nl label. In experiments, we found that the second approach performs slightly better and thus we report the results based on this approach.
Nevertheless, our results show that querying the proposed classifier with adversarial examples does not help much to train a better substitute classifier, since the target classifier returns \nl for most of the queries and thus the substitute classifier gains little information about the target's decision boundaries.

We examine the transferability rates of the adversarial examples generated using $\mathtt{Grad_0}$ and FGS methods (for misclassification attack) and the greedy method (for targeted attack), under both black-box and blind models. 
{\it For the MNIST dataset, in all cases the transferability rate is zero.} 
This means that MNIST data have enough redundancy so that the classifier can perfectly distinguish between the clean data and adversarial examples transferred from other classifiers. 
However, for the GTSRB dataset, the classifier cannot discard all the adversarial examples while keeping the clean samples. Figure \ref{fig:attack_null} shows the transferability rate of the adversarial examples. The classifier typically maps the adversarial examples with small perturbations to their original label, and the adversarial examples with high perturbations to the \nl label. However, moderate perturbations can cause the classifier to mistakenly label the input. Nevertheless, the transferability rate is near zero for adversarial examples generated with $L_0$ norm constraint and less than $10\%$ with $L_{\infty}$ norm constraint.

\subsection{Discussion}
Here, we discuss the advantages of enhancing ML classifiers with an additional label to detect out-of-distribution inputs.

\vspace{0.1cm}
\noindent{\it The Essence of \nl Labeling.} 
During the training process, the classifier gradually learns to properly place the data into the multi-dimensional space of the input features and form the class boundaries by packing the data of different classes into clusters. However, for most of naturally distributed data, most of the space is empty; this is verified by the observations that projecting the space down to even a relatively small number of principal components usually yields reasonable accuracy~\cite{abdi2010principal}. 
Therefore, the classifier is never trained for the most of the input space, because of the implicit assumption that it will be always tested on the data that follow the same distribution as the training data. This however exposes the classifier to maliciously perturbed or out-of-distribution inputs.

As a result, we need a mechanism for classifiers to reject the invalid inputs~\cite{amodei2016concrete}. We hypothesize that ML classifiers, aside from learning the boundaries of different classes, can be trained to also learn the distribution of the valid input data. Hence, a classifier can first inspect the distribution of an input, and if it is not close to that of the training data, it will be rejected. In this way, we can practically block the {\it unused} space of the input features and robustify the classifier to adversarial examples. To implement this idea, we propose adding an extra label, called \nl label, and train the classifier to recognize and reject the out-of-distribution inputs by classifying them as \nl.

\vspace{0.1cm}
\noindent{\it Human-in-the-loop AI.} 
When receiving an adversarial example, one approach is to determine the original label of the input. In this paper, we suggest that an alternative approach would be to reject the input and potentially obtain another measurement or ask a human expert to inspect the data. This approach proves to be safer and more reasonable in many applications, as discarding an input may incur only a small cost, whereas, misclassifying or even accepting an adversarial sample can lead to a harmful situation.
For example, in applications such as medical diagnosis and bioinformatics, the cost of rejecting a sample can be additional medical tests, which is far more acceptable than taking the risk of deciding based on a potentially perturbed sample. As another example, consider self-driving cars, for which ML classifiers are used to recognize signs or other vehicles on the road~\cite{cirecsan2012multi}. Perturbing the input of such systems may force the classifier to misclassify it and may eventually cause an accident. In contrast, if the classifier detects that the input is untrusted, it can quickly notify the human deriver to take over until the threat is resolved.

Rejecting the dubious data in the test phase has been implemented in the ``human-in-the-loop'' computing. The idea is that if the classifier's confidence in the output label is below a certain value, it sends the data to a human annotator to make a judgment~\cite{AI}. 
Adversarial examples are however subtly modified data which can cause the classifier to report high confidence in a wrong class. Therefore, the current {\it rejection} systems are not resistant against such adversaries. Instead, we proposed the \nl labeling method to systematically train the classifier to detect anomalies, rather than just discarding the data based on the confidence value. Therefore, our proposed training method can further empower the current systems to defend against broader range of adversaries.

\section{Related Work}\label{related}

The security of learning systems has been studied from different perspectives~\cite{barreno2006can,barreno2010security,huang2011adversarial,biggio2014security,amodei2016concrete,papernot2016towards}. One vulnerability of ML classifiers is the existence of adversarial examples~\cite{szegedy2013intriguing}. 
In this section, we review the works on the related attack and defense methods.

\subsection{Attacks Methods}
Several works have proposed methods for generating adversarial examples on ML classifiers~\cite{goodfellow2014explaining,papernot2016limitations,carlini2016towards,moosavi2016deepfool}.
In~\cite{tabacof2016adversarial,kos2017adversarial}, adversarial examples have been also applied to other ML models, such as variational autoencoder~\cite{vincent2010stacked} and generative adversarial networks~\cite{goodfellow2014generative}. 
A related concept to adversarial examples is {\em fooling images}, which are images that are completely unrecognizable to humans, but ML classifiers confidently map them to recognizable objects~\cite{nguyen2015deep}.

Preceding papers demonstrated the vulnerability of learning systems against adversaries with full knowledge about the classifier. Another line of works assumes that the adversary does not have access to the classifier parameters. The attacks on black-box learning systems rely on the transferability property of the adversarial examples~\cite{szegedy2013intriguing}. This property has been exploited in~\cite{papernot2016practical,papernot2016transferability} to mount attacks on ML oracles by transferring adversarial examples from a substitute classifier. In~\cite{liu2016delving}, the authors proposed ensemble-based approaches to generate transferable adversarial examples. 
In~\cite{moosavi2016universal}, the authors showed the existence of a universal (image-agnostic) adversarial perturbation that transfers across images as well as different classifiers.


In this paper, we considered a black-box model where the adversary possesses a small dataset and has an oracle access to the target, i.e., she can obtain the label for any input by querying the target classifier. We also proposed the blind model where the adversary only possesses a small dataset and has no access to the target classifier. We showed that we can achieve high transferability rates for both misclassification and targeted attacks, even when a defense method has been employed.
Similar to many of the previous works, e.g., \cite{szegedy2013intriguing,goodfellow2014explaining,papernot2016limitations,papernot2015distillation,papernot2016practical,papernot2016transferability,kurakin2016adversarial,carlini2016towards}, we demonstrated the results on image classifiers, though adversarial examples can be  generated also for other data types~\cite{grosse2016adversarial,kereliuk2015deep,reddy2016obfuscating,hosseini2017deceiving}.

\subsection{Defense Methods}
Existing defense techniques can be grouped into four categories of preprocessing methods, methods based on regularization and adversarial training, the distillation method and classification with rejection. In the following, we review the defense methods.

\vspace{0.1cm}
\noindent {\bf Preprocessing Methods.} 
Natural images have special properties, such as high correlation among adjacent pixels, sparsity in transform domain or having low energy in high frequencies~\cite{bovik2010handbook}. Hypothesizing that adversarial changes do not lie in the same space as natural images, several works considered reversing the adversary's effect by passing the images through a filter~\cite{maharajimproving,dziugaite2016study,wang2016random,wang2016using,graese2016assessing}.
Although preprocessing the inputs makes the attack more challenging, it does not remove its possibility. Moreover, these filters usually decrease the classifier accuracy on clean data. 


\vspace{0.1cm}
\noindent {\bf Regularization and Adversarial Training.} 
Several works proposed improving the robustness of ML classifiers by regularization~\cite{gu2014towards,zhao2016suppressing,rozsa2016towards}, adversarial training~\cite{goodfellow2014explaining,lyu2015unified,shaham2015understanding,huang2015learning,miyato2015distributional,nokland2015improving,demyanov2015invariant}, or label smoothing~\cite{warde2016adversarial}. In these methods, the accuracy on adversarial examples is however substantially lower than the accuracy on clean data. For instance, in one experiment in~\cite{goodfellow2014explaining}, it is reported that without adversarial training, the classifier had an error rate of $89.4\%$ on adversarial examples and, with adversarial training, the error rate fell to $17.9\%$; however, the error rate of clean data is reported as smaller than $1\%$. This means that we cannot fully rely on these defense mechanisms when deploying learning systems in critical applications. 
Moreover, in experiments we showed that we can successfully mount attacks on robust learning systems under the blind model. Therefore, these methods are not effective against adversarial examples.

\vspace{0.1cm}
\noindent {\bf Distillation Method.} 
In~\cite{papernot2015distillation}, Papernot et al. proposed a mechanism called defensive distillation to reduce the effectiveness of adversarial samples, by preventing the attacker from progressively selecting adversarial features. In~\cite{carlini2016towards}, the authors showed that an adversary, with access to the classifier parameters, can revert the effect of the distillation. We have studied the distillation method also in the black-box and blind models and verified that it does not improve the classifier's robustness. The reason is that although the distillation method significantly reduces the gradient of the loss function with respect to small input changes, it is not effective when a set of the features are modified at once, which is the case in the black-box and blind models.

\vspace{0.1cm}
\noindent{\bf Classification with Rejection.}
A line of work studied the problem of classification with rejection, where the system can choose not to classify an observation~\cite{elkan2001foundations,fumera2000reject,herbei2006classification,bartlett2008classification,cortes2016learning}. The reject option is used, for instance, when the conditional class probabilities are close. 
In a related work, in~\cite{hendrycks2016baseline}, the authors observed that correctly classified examples tend to have greater maximum probabilities than erroneously classified and out-of-distribution examples. 
However, it has been shown that adversarial examples can result in high confidences in the wrong classes~\cite{goodfellow2014explaining}. Therefore, rejecting the input by examining the corresponding output probability distribution is not effective in detecting the adversarial examples. 

A similar approach is classifying invalid data as a ``rubbish class'' \cite{bromley1993improving,lecun1998gradient,yadav2014novelty}. In this approach, the classifier is trained to assign a uniform distribution to invalid data and then discards the samples with low confidence during the test time. We have evaluated this method and verified that, while it improves the robustness of the classifier to adversarial examples, it does not provide similar trade-offs to our proposed method for the test accuracy and detection rate of the adversarial examples. 
In~\cite{metzen2017detecting}, the authors proposed augmenting the classifier with a detector subnetwork which is a binary classifier for distinguishing the clean data from adversarial examples; the authors however did not provide the false alarm rate, i.e., the rate of wrongly detecting clean data as adversarial. Moreover, it was noted that an adversary can generate adversarial examples that fool both the classifier and the detector. 

\vspace{0.1cm}
In this paper, we proposed a training method that imitates the human reasoning by design; we augmented the output class set of the classifier with an additional label and adaptively train the classifier to assign adversarial examples to the new label with proper probability. Our method does not increase the complexity of deploying ML systems in real-world settings. We evaluated our method in black-box models and demonstrated that it effectively detects the adversarial examples, without wrongly flagging the clean test data.


\section{Conclusion}

In this paper, we studied the problem of robustifying black-box learning systems against adversarial examples. Adversarial examples are inputs, which are maliciously perturbed in order to deceive the learning system into providing erroneous outputs. Adversarial examples are known to transfer between classifiers; therefore, an adversary can attack a black-box learning system, by generating the adversarial example on a substitute classifier and then transfer them to the target classifier. We performed extensive experiments on different ML classifiers and demonstrated their vulnerability against adversarial examples even when a defense mechanism is employed. 

We then proposed the concept of \emph{\nl labeling}, where we intend to identify and discard adversarial examples, instead of classifying them into their original label. Our approach addresses a fundamental problem of current ML classifiers that, while only a tiny fraction of the input space is used for training, all possible inputs are allowed during the test phase. This causes the classifier to show unexpected behaviors when it is queried with invalid inputs. In contrast, we train the classifier to distinguish between the clean and adversarial data. Our experimental results show that the proposed method effectively blocks the transferability of the adversarial examples by mapping them to the \nl label, while classifying the clean test data into the valid labels. 

\vspace{0.4cm}
\noindent{\bf \large Acknowledgments}

\vspace{0.1cm}
\noindent This work was supported by ONR grants N00014-14-1-0029 and N00014-16-1-2710, ARO grant W911NF-16-1-0485 and NSF grant CNS-1446866.



%

\bibliographystyle{ieeetr}
\bibliography{Main}

\appendix
\section{DNNs Used for Training}\label{apx:DNNs}

We demonstrated the results on deep neural networks trained on the MNIST and GTSRB datasets. For both datasets, the networks have two convolutional layers with $32$ and $64$ filters and with a kernel size of $3\times 3$, followed by a max-pooling layer. 
For MNIST, the convolutional layers are followed by two fully-connected layers with $200$ rectified linear units. For GTSRB, the convolutional layers are followed by two fully-connected layers with $512$ and $128$ rectified linear units, with a dropout rate of $0.5$. In both networks, the classification is made by a softmax layer.

\section{Methods for Generating Adversarial Examples}\label{apx:Adv_gen}

In this section, we explain the methods used in the paper for generating the adversarial examples. 

Let $\ell(X, y; \theta)$ denote the loss of the classifier with parameters $\theta$ on $(X, y)$. The loss function quantifies the classifier's goodness-of-fit for the sample $(X, y)$ under the parameters $\theta$. When holding $\theta$ fixed and viewing the loss as a function of $(X,y)$, we simply write $\ell(X, y)$. We alternatively write $\ell(X, Z)$ for the loss of the classifier for input $X$ and output probability vector of $Z$. 

\subsection{Adversarial Examples for Misclassification Attack}

In misclassification attack, the adversary intends to find a perturbation which causes the loss function to increase the most. Therefore, we have:
\begin{align}\label{eq:ell}
\delta X^*=\argmax_{\|\delta X\| \leq \epsilon \cdot \|X\|}\ell(X+\delta X, y).
\end{align}
In general, solving~\ref{eq:ell} is difficult due to its non-convex nature with respect to $\delta X$ and $\theta$. Several works proposed solving the problem by linearizing the loss function at the point $X$. Therefore, we can write Equ.~\ref{eq:ell} as follows:
\begin{align}\label{eq:delta_x}
\nonumber \delta X^*=&\argmax_{\|\delta X\| \leq \epsilon \cdot \|X\|}\ell(X+\delta X,y)\\
\nonumber \approx&\argmax_{\|\delta X\| \leq \epsilon \cdot \|X\|} \ell(X,y)+\nabla_X \ell(X,y)^T \delta X \\
=&\argmax_{\|\delta X\| \leq \epsilon \cdot \|X\|} \nabla_X \ell(X,y)^T \delta X
\end{align}
where $\nabla_X \ell(X,y)$ is the gradient of the function $\ell$ with respect to the vector $X$.
Based on this method, in~\cite{goodfellow2014explaining}, Goodfellow et al. proposed Fast Gradient Sign (FGS) method for generating adversarial examples with $L_{\infty}$ constraint. In this paper, we adopted this method for the $L_0$ norm constraint, i.e., we generate adversarial examples with changing few input features, called {\it adversarial features}, and proposed two methods, namely $\mathtt{Grad_0}$ and MG methods. In the following, we provide the solutions for $\mathtt{Grad_0}$, MG and FGS methods.

\vspace{0.2cm}
\noindent{\bf $\mathtt{Grad_0}$ Method.} 
In this method, the adversary maximizes the term $\nabla_X \ell(X,y)^T (X^*-X)$ such that $\|\delta X\|_0 = \epsilon \cdot |X|$. Therefore, to compute $X^*$, we do the following procedure:
\begin{itemize}
	\item Compute $\nabla_X \ell(X,y)$ and define a binary vector $X'$, for which $X'_i=1$ if and only if $\nabla_X \ell(X,y)_i > 0$.
	\item Let $S$ be the set of indexes of $\epsilon \cdot |X|$ largest elements of $\nabla_X \ell(X,y) \circ (X'-X)$, where $\circ$ denotes the element-wise product. 
	\item Obtain $X^*_S=X'_S$ and $X^*_{\bar{S}}=X_{\bar{S}}$, where $\bar{S}$ denotes the complement set of $S$, and $X_S=\{X_i | i\in S\}$. 
	\item Return $X^*$ as the adversarial example, if the label of $X^*$ is different from the label of $X$; if the label has not changed, declare infeasibility.
\end{itemize}

\vspace{0.2cm}
\noindent{\bf MG Method.}
This method is similar to the $\mathtt{Grad_0}$ Method in that the adversary maximizes $\nabla_X \ell(X,y)^T (X^*-X)$. We however take a greedy approach, i.e., we select and modify one feature at a time until the input is successfully misclassified.


\vspace{0.2cm}
\noindent{\bf $\mathtt{Grad_{\infty}}$ Method.} 
The solution to this method is provided in~\cite{goodfellow2014explaining} as the following:
$$X^*=X+\epsilon \mbox{sign}{\nabla_X \ell(X,y)}.$$
We return $X^*$ as the adversarial example, if the label of $X^*$ is different from the label of $X$; if the label has not changed, we declare infeasibility.

\subsection{Adversarial Examples for Targeted Attack}

In targeted attack, the adversary intends to find a perturbation that minimizes the loss function for the input $X$ and target label $y^*$ or target vector $Z^*$. Similar to Equations~\ref{eq:ell} and~\ref{eq:delta_x}, we write:
\begin{align}\label{eq:delta_x_target}
\nonumber \delta X^*=&\argmin_{\|\delta X\| \leq \epsilon \cdot \|X\|}\ell(X+\delta X,y^*)\\
\approx&\argmin_{\|\delta X\| \leq \epsilon \cdot \|X\|} \nabla_X \ell(X,y^*)^T \delta X.
\end{align}

In this paper, based on Equ.~\ref{eq:delta_x_target}, we proposed two methods for generating adversarial examples for trageted attack, namely Greedy and STG Methods. In the following, we provide the solutions for these methods.

\vspace{0.2cm}
\noindent{\bf Greedy Method.}
This method is similar to the MG method in that the adversary takes a greedy approach for finding the adversarial perturbation. However, in this method, at each iteration we select and modify the feature that minimizes $\nabla_X \ell(X,y^*)^T (X^*-X)$. Note that the proposed greedy method is significantly faster than the current greedy approaches, e.g.~\cite{papernot2016limitations,carlini2016towards}, because for selecting each feature, we only do one forward run and one blackpropagation; while current methods need to run the network at least $|X|$ number of times, for selecting each feature.

\vspace{0.2cm}
\noindent{\bf STG Method.}
In this method, we set the target output probability vector $Z_T$ to be $\frac{1}{K}$ for the $K$ valid classes and use the $L_0$ norm constraint for generating adversarial examples. The adversary thus minimizes $\nabla_X \ell(X,Z_T)^T (X^*-X)$ such that $\|\delta X\|_0 = \epsilon \cdot |X|$. Similar to the $\mathtt{Grad_0}$ method, to compute $X^*$, we do the following procedure:
\begin{itemize}
	\item Compute $\nabla_X \ell(X,Z_T)$ and define a binary vector $X'$, for which $X'_i=1$ if and only if $\nabla_X \ell(X,Z_T)_i < 0$.
	\item Let $S$ be the set of indexes of $\epsilon \cdot |X|$ smallest elements of $\nabla_X \ell(X,Z_T) \circ (X'-X)$, where $\circ$ denotes the element-wise product. 
	\item Obtain $X^*_S=X'_S$ and $X^*_{\bar{S}}=X_{\bar{S}}$, where $\bar{S}$ denotes the complement set of $S$, and $X_S=\{X_i | i\in S\}$. 
	\item Return $X^*$.
\end{itemize}

\end{document}